\pdfoutput=1
\documentclass{article}
\usepackage{arxiv}            % placeholder preprint style; switch to the NeurIPS 2027 Datasets & Benchmarks kit when released
\usepackage[utf8]{inputenc}
\usepackage[T1]{fontenc}
\usepackage{microtype}
\usepackage[round]{natbib}
\usepackage{amsmath,amsfonts,bm}

\def\eqref#1{equation~\ref{#1}}
\def\1{\bm{1}}

\DeclareMathAlphabet{\mathsfit}{\encodingdefault}{\sfdefault}{m}{sl}
\SetMathAlphabet{\mathsfit}{bold}{\encodingdefault}{\sfdefault}{bx}{n}

\usepackage{hyperref}
\usepackage{url}
\usepackage{booktabs}
\usepackage{array}
\usepackage{graphicx}
\usepackage{amsmath,amssymb}
\usepackage{xcolor}
\usepackage{multirow}
\usepackage{tikz}
\usetikzlibrary{arrows.meta,positioning,fit,calc}
\definecolor{cState}{RGB}{31,97,175}
\definecolor{cCall}{RGB}{34,139,94}
\definecolor{cRule}{RGB}{214,110,20}

\newcommand{\cls}[1]{\texttt{#1}}
\newcommand{\bench}{\textsc{Careful}-Bench}   % CAREFUL: Confirmation-Aware Replay Evaluation of Function calling Under effect Labels

\newcommand{\nLeaderboardBase}{11}
\newcommand{\nLeaderboardFull}{11}
\newcommand{\spearmanBase}{0.98}
\newcommand{\spearmanMean}{1.00}
\newcommand{\largestDrop}{14.5}

\newcommand{\surplusOutMin}{3}
\newcommand{\surplusOutMax}{44}
\newcommand{\surplusInMin}{2}
\newcommand{\surplusInMax}{28}
\newcommand{\budgetHaikuRun}{3}

\newcommand{\budgetSonnetRun}{54}
\newcommand{\budgetSonnetAll}{161}
\newcommand{\budgetOpusRun}{87}
\newcommand{\budgetOpusAll}{262}
\newcommand{\budgetFableRun}{193}
\newcommand{\budgetFableAll}{580}
\newcommand{\budgetAllModels}{1,010}
\newcommand{\apiInOutRatio}{68}
\newcommand{\apiConversations}{800}
\newcommand{\apiPromptTokensM}{107}
\newcommand{\apiOutputTokensM}{1.6}
\newcommand{\apiCacheShare}{93}
\newcommand{\apiRequests}{8,696}
\newcommand{\haikuSpent}{2.75}
\newcommand{\haikuSpentCons}{12.21}
\title{CAREFUL-Bench: Effect-Aware, Judge-Free Evaluation\\of Multi-Turn LLM Function Calling}
\renewcommand{\headeright}{Preprint}
\renewcommand{\shorttitle}{CAREFUL-Bench: effect-aware, judge-free evaluation of function calling}
\renewcommand{\undertitle}{Preprint}
\date{\today}

\author{
Alexander~Kovalenko\thanks{Corresponding author.} \\
Faculty of Information Technology \\
Czech Technical University in Prague \\
\And
Artem~Redchych \\
Faculty of Information Technology \\
Czech Technical University in Prague
}

\begin{document}
\maketitle

\begin{abstract}
The multi-turn checker of the Berkeley Function Calling Leaderboard (BFCL) checks only the final state of a simulated
service, never the calls that produced it. Extra calls cost nothing, no function doc says what a function does, and the
scripted user never answers a proposal, so a model that asks before deleting or paying fails the turn; 322 of the 734
reference turns in the base category act destructively without confirmation. We add effect information to BFCL without
replacing it. Effect declarations for all 128 functions are inferred by executing each function in BFCL's own simulators
and diffing the state (read or write, idempotent or not, footprint); the remaining judgment, which writes are
destructive, is made by four annotators, two humans and two language models (Fleiss' $\kappa=0.75$), and adjudicated. A judge-free replay scorer re-scores any logged BFCL
run, reproduces BFCL's results on all 800 reference trajectories, and reports, alongside the official score, unrequested
destructive calls, duplicate writes, redundant reads, surplus steps and their token cost. A confirmation-aware protocol
adds a deterministic user that answers proposals. On \nLeaderboardBase{} open models the effect-aware score preserves
the official ranking ($\rho=\spearmanBase$) but changes the gaps: the leader loses \largestDrop{} points to unrequested
destructive calls, and surplus steps account for \surplusInMin--\surplusInMax\% of all input tokens. Effect hints in the
function docs produce no measurable change for any of eight models, consistent with their advisory status in the Model
Context Protocol. A user that answers recovers up to four points for the models that ask and none for the rest, with
every comparison paired on the same conversations. The gains come from clarifying questions rather than requests for
permission: of the 712 conversations that require a destructive call, no model completes more than 6 with every such
call confirmed, a difference the official score does not record.
\end{abstract}

% =====================================================================================================================
\section{Introduction}
\label{sec:intro}

A function-calling benchmark gives a model a set of function docs and a user request, lets the model call the functions
against a simulated service, and checks the outcome. The checks have become more realistic over three generations. The
first compared the abstract syntax tree of a predicted call with a reference \citep{patil2024gorilla}; the second
executed the call; the third compares the \emph{state} that a sequence of calls leaves behind in a stateful service
\citep{patil2025bfcl}. None of these checks examines the calls themselves. A model that reaches the right state after
deleting and recreating a file, posting a tweet twice, or reading the same inbox four times receives the same score as
one that made only the reference calls. A model that stops before an irreversible call to ask the user receives a
lower score, because the user is a script and does not answer.

The behaviour that these checks miss is the behaviour practitioners increasingly ask of tool-using agents. The Model
Context Protocol lets a tool be annotated as read-only, destructive or idempotent \citep{mcp2025annotations}. Policy
benchmarks require confirmation before a cancellation \citep{yao2024taubench,barres2025tau2}. Runtime guards block or
gate calls by their effect \citep{kang2026polar,nexus2026}. The benchmark on which function calling is ranked, however,
treats every call alike. In previous work on effect-typed training data \citep{kovalenko2026effectworld} we found that
an adapter trained to confirm before destructive calls lost about 1.5 points of BFCL multi-turn accuracy for doing so.
The same leaderboard did not register the two- to four-fold increase in extra write calls that the same adapters made on
BFCL's services.

We extend BFCL rather than replace it. We work on its multi-turn categories as released (\cls{bfcl-eval} 2026.3.23),
because that is the tool set the community already evaluates on, and add a layer we call \bench{}, for
Confirmation-Aware Replay Evaluation of Function calling Under effect Labels. It has five parts.
\begin{enumerate}\setlength{\itemsep}{0pt}
\item \textbf{Effect declarations for all 128 multi-turn functions}, inferred empirically (\S\ref{sec:effects}). We execute
  each function in BFCL's own simulators from sampled states and diff the state. A function that changes nothing is a
  read and one that changes something is a write; a write is idempotent if applying it again leaves the state as it is;
  the attributes it changes are its footprint. Whether a write is \emph{destructive} (deletes data, reaches other people
  or moves money) is a judgment rather than an observation. Three consequence signals computed from the diffs propose
  candidates, and four annotators (two humans and two language models, each blind to the others) label the boundary.
  We report their agreement and take the majority.
\item \textbf{Effect-aware scoring by replay} (\S\ref{sec:scoring}). Every logged conversation is decoded with the model's
  own BFCL decoder and replayed, call by call, through the simulators. Deterministic rules then count destructive calls
  that the reference does not make, duplicate non-idempotent writes, redundant reads, failed calls and calls beyond the
  reference. A conversation succeeds \emph{under effect safety} only if it passes BFCL's checker and every rule. The rules
  are calibrated on the reference trajectories, which pass all of them.
\item \textbf{A confirmation-aware protocol} (\S\ref{sec:protocol}). A deterministic scripted user answers proposals for
  destructive calls (yes, or no with the original request restated), guarded by a strict question detector, so that
  asking before acting can be rewarded rather than penalised.
\item \textbf{An effect-hints condition} (\S\ref{sec:experiments}). The same tasks are run with the declarations rendered
  into the function docs in the vocabulary of the Model Context Protocol, to test whether models use effect information
  when it is offered.
\item \textbf{A leaderboard} of open models across families and sizes, and of API models within a fixed budget, under the
  standard and the effect-aware protocol (\S\ref{sec:experiments}: \nLeaderboardBase{} open models and two API models under
  the standard protocol; seven open models and the API model under effect hints and under confirmation).
\end{enumerate}
No language model grades another at any stage. The declarations, the scorer and the protocol are released as a thin layer
over the pinned \cls{bfcl-eval} package under the same Apache-2.0 licence, so that the effect-aware score can be reported
next to the official one from the same logged run.

% =====================================================================================================================
\section{Related work}
\label{sec:related}

\paragraph{State-based function-calling benchmarks.} BFCL \citep{patil2025bfcl} is our starting point. Its multi-turn
categories run the model against eight simulated services (a file system, messaging, a social network, a ticket system,
a trading account, travel booking, vehicle control and a calculator) and check, after each user turn, that the model's
instances have the same public state as the reference's and that the reference's function results appear among the
model's. ToolSandbox \citep{lu2025toolsandbox} adds implicit state dependencies and milestone scoring along the
trajectory. AppWorld \citep{trivedi2024appworld} grades the final database state of nine applications with unit tests
that also catch unintended changes. NESTful \citep{basu2024nestful} tests nested call sequences. WorkBench
\citep{styles2024workbench} flags a side effect when the agent's database changes differ from the ground truth, and
AgentRewardBench \citep{lu2025agentrewardbench} has experts mark unintended side effects in web-agent trajectories; both
record side effects as a single flag per episode. None of these benchmarks declares what a function does or counts calls
that were unnecessary, repeated or unconfirmed. \citet{bellibatlu2026audit} audit the mutating tools of four agent
benchmarks against their advertised interfaces and trace scores back through evaluator code. Our \S\ref{sec:bfcl} is an
audit of the same kind for BFCL's checker, and our declarations supply the interface contract that its tools currently
lack.

\paragraph{Policy and consent.} $\tau$-bench and $\tau^2$-bench \citep{yao2024taubench,barres2025tau2} score an agent
against written domain policies in conversation with a language-model user, so that confirming before a cancellation is
part of the task. The user is non-deterministic, however, and the domains are few. ST-WebAgentBench
\citep{levy2026stwebagentbench} pairs web tasks with policies, user consent among them, and credits only completion under
policy; our success under effect safety applies the same idea to function calling. ClawsBench \citep{li2026clawsbench}
runs productivity agents against mock mail, calendar and document services and finds unsafe actions in a sizeable share
of runs, but with a small task set and no effect declarations. When2Call \citep{ross2025when2call} and ToolBeHonest
\citep{zhang2024toolbehonest} test the decision to call, ask or decline, without state or effects.

\paragraph{Risk by emulation or judgment.} ToolEmu \citep{ruan2024toolemu} emulates tools with a language model to
surface risky actions, R-Judge \citep{yuan2024rjudge} asks whether a model recognises a risky trajectory, and AgentHarm
\citep{andriushchenko2025agentharm} measures compliance with harmful tasks. All three need a judge or an emulator, and
all three measure harm rather than effect discipline on benign tasks. Our rules are evaluated by replay in the
benchmark's own simulators, so every count is reproducible at the level of individual calls.

\paragraph{Effects at runtime.} Several 2026 systems classify tool effects in order to guard execution. Atomix
\citep{mohammadi2026atomix} records reads and effects to give tool use transactional semantics; NEXUS \citep{nexus2026}
decides per call whether to allow, block, ask for confirmation or ask for revision; TraceGrant \citep{tracegrant2026}
issues per-call certificates from a contract compiled from the request; POLAR \citep{kang2026polar} blocks
hard-to-reverse actions; \citet{doshi2026verifiably} enforce specifications over capability-labelled tools; and
\citet{odersky2026capabilities} have agents write code whose capabilities a type system tracks. All of these presuppose
effect information. None measures whether models behave well with or without it; that is the question this benchmark
asks. Contract2Tool \citep{contract2tool2026} is the method closest to our inference. It learns preconditions and
effects from docs and before/after traces and uses them for tool filtering rather than scoring, whereas we infer from
controlled executions in the simulators themselves and use the result to score. The Model Context Protocol's tool
annotations \citep{mcp2025annotations} are the practical interface for such information. The specification calls them
advisory, and a census of registered tools finds them widely emitted but coarse \citep{trofimov2026boundary}; our hints
condition can be read as an evaluation of that design. Repeated writes have been measured in agents outside benchmarks
\citep{li2026exactlyonce}; our duplicate-write rule counts them inside one. Table~\ref{tab:related} places the benchmark
among these neighbours by five properties: whether the tools carry effect declarations, whether scoring is deterministic
(judge-free), whether the user can answer a proposal, whether extra calls are scored (a flag marks a side-effect flag
only), and whether the tool set is already in common use.

\begin{table}[t]
\centering\small\setlength{\tabcolsep}{4pt}
\caption{Benchmark comparison.}
\label{tab:related}
\begin{tabular}{lccccc}
\toprule
 & effects & judge-free & answers proposals & extra calls & common tool set \\
\midrule
BFCL v4 multi-turn \citep{patil2025bfcl} & -- & \checkmark & -- & -- & \checkmark \\
$\tau$ / $\tau^2$-bench & -- (policies) & -- (LM user) & \checkmark & -- & \checkmark \\
ST-WebAgentBench & -- (policies) & \checkmark & \checkmark & -- & web \\
ClawsBench & -- & partly & -- & -- & -- \\
WorkBench / AgentRewardBench & -- & \checkmark / -- & -- & flag & -- \\
ToolEmu / R-Judge / AgentHarm & -- & -- & -- & -- & -- \\
Runtime guards (Atomix, NEXUS, POLAR) & \checkmark & n/a & n/a & n/a & n/a \\
\bench{} (this paper) & \checkmark (inferred) & \checkmark & \checkmark & \checkmark & \checkmark (BFCL) \\
\bottomrule
\end{tabular}
\end{table}

% =====================================================================================================================
\section{BFCL's multi-turn checker}
\label{sec:bfcl}

We read the checker of \cls{bfcl-eval} 2026.3.23 and measured its data. The four multi-turn categories (\cls{base},
\cls{miss\_func}, \cls{miss\_param}, \cls{long\_context}) share 200 conversations over eight service classes
(\cls{GorillaFileSystem}, \cls{TradingBot}, \cls{TravelAPI} and \cls{VehicleControlAPI} in 50 conversations each,
\cls{MessageAPI} 40, \cls{TwitterAPI} 39, \cls{TicketAPI} 31, \cls{MathAPI} 25); the classes expose 128 public methods,
each with a function doc. A conversation has several user turns; for each turn a reference trajectory lists the calls
that solve it.

\paragraph{How a turn is checked.} The model's calls in a turn are decoded with the model's own decoder (steps that do
not decode are dropped silently) and executed in order on instances created from the conversation's initial
configuration; the reference calls are executed on a second set of instances. The turn passes if (i) every instance has
the same \emph{public} attributes as its reference twin, where attributes with a leading underscore are skipped, and (ii)
every function result of the reference turn appears among the model's results so far, as an unordered sub-multiset. A
turn with reference calls but no decoded model call fails, and a conversation that exceeds the step limit fails. A check
of the order of method invocations exists in the source but is disabled. Three properties of this check matter for what
follows. Extra calls are free, since any call that leaves the public state as the reference does and does not prevent
the reference results from appearing costs nothing. Hidden state is invisible; the file system's current directory, for
example, is a private attribute, so a wrong \cls{cd} goes unnoticed unless a later call exposes it. The user never
answers, because user turns are scripted in advance, so a turn in which the model asks a question and waits is a turn
with no call, and such a turn fails whenever the reference calls a function.

\paragraph{The reference trajectories.} Under the declarations of \S\ref{sec:effects}, the 1,142 reference calls of
\cls{multi\_turn\_base} split into 342 destructive calls, 268 other writes and 532 reads. Of the 734 reference turns, 322
make at least one destructive call (deleting files, sending messages, posting, placing or cancelling orders, booking or
cancelling flights), and none of these is preceded by a confirmation; 432 turns make some write, and only 3 turns expect
no call. The \cls{miss\_func} and \cls{miss\_param} categories add one turn per conversation (934 turns in total) in
which the model should \emph{not} call because a function or a parameter is missing, so 203 of their turns reward asking
for information. No turn in any category rewards asking for permission. The references use 81 of the 128 functions. The
remaining 47 are documented and callable but never part of a solution, which matters for \S\ref{sec:effects} because
their effects cannot be inferred from the references alone. The function docs describe arguments and return values and
say nothing about effects; none of the 128 descriptions tells the model that a call deletes, sends or pays.

% =====================================================================================================================
\section{Effect declarations by state diff}
\label{sec:effects}

An effect declaration for a function says whether a call reads or writes, whether a write is idempotent, which part of
the state it touches, and whether it is destructive. Writing 128 such declarations by hand from the source is possible but
slow and, as we found in the teaser table of \citet{kovalenko2026effectworld}, error-prone; there, hand labels classified
the hidden-state \cls{cd} as a write and \cls{mv} as destructive. We therefore infer everything that can be inferred by
executing the functions and reserve human judgment for the destructive boundary alone.

\paragraph{Observations.} An observation is one call executed on a simulator instance in a known state, with a snapshot of
the state before and after. Snapshots contain the public attributes BFCL compares plus the private ones, minus four
artefacts (random-number generators, a wall-clock stamp, a description string, and a price-quote memo of the travel
service that \cls{get\_flight\_cost} fills and \cls{book\_flight} consumes). We take observations from three sources.
The first is the 800 reference trajectories of the four categories. The second is 4,388 logged conversations of six
models from \citet{kovalenko2026effectworld}, decoded with their own decoders, which add realistic mistakes and error
paths. The third is synthetic probes, in which each function is called from up to 40 sampled states per class (the
initial state and the state after each reference turn), with argument sets taken from reference calls of that function
where they exist and synthesised from the doc's parameter types otherwise, until 20 successful calls are collected. In
total 5,188 traces and 2,041 successful probe calls cover every function; 464 decoded calls in the logs named functions
that do not exist.

\paragraph{Inference.} A call is \emph{successful} unless it raised, returned an error field, or returned a false status
flag (the simulators' soft failures, such as unfollowing someone not followed or logging out when logged out). A function
is a \textbf{write} if any successful call changed the state and a \textbf{read} otherwise. A write is
\textbf{idempotent} if, in every probe, applying the same call again to the resulting state changed nothing; since the
probes sample states rather than enumerate them, this is evidence of idempotence rather than a proof. The
\textbf{footprint} of a write is the set of attributes its successful calls changed, with entity keys generalised. One
function, \cls{get\_outside\_temperature\_from\_weather\_com}, returns an error by design; it never changed the state in
82 failing calls and is declared a read on that evidence. The result is 79 reads and 49 writes, of which 16 are non-idempotent and one, \cls{cd}, changes only hidden state, which
BFCL's checker never compares; Table~\ref{tab:effects} gives the counts per simulator class, with the destructive
boundary of the next paragraph. Several outcomes differ from what
the docs suggest. \cls{rm}, \cls{mkdir}, \cls{retweet} and \cls{cancel\_order} are idempotent because the second call
fails harmlessly; \cls{cd} into a named child is idempotent but \cls{cd ..} is not; and \cls{cp} is not idempotent.

\paragraph{The destructive boundary.} Following \citet{kovalenko2026effectworld}, a write is \emph{destructive} if its
effect reaches a third party or cannot be undone: it deletes or overwrites data, reaches other people, moves money, or
acts in the physical world. Three signals computed from the diffs propose candidates: entries removed (\cls{rm},
\cls{rmdir}, \cls{mv}, \cls{delete\_message}, \cls{remove\_stock\_from\_watchlist}, \cls{cancel\_booking}), a balance
changed (\cls{fund\_account}, \cls{withdraw\_funds}, \cls{book\_flight}, \cls{purchase\_insurance},
\cls{cancel\_booking}), and an entry added to another person's inbox, timeline or thread (\cls{send\_message},
\cls{post\_tweet}, \cls{comment}, \cls{mention}, \cls{retweet}). Four annotators then labelled all 128 functions, each
blind to the others: A1, coded by the first author from the signals and the simulator source; A2 and A3, the language
models Claude Opus 5.5 and Claude Fable 5.1, from the review sheet and the simulator source; and A4, the second author,
from a copy of the review sheet with the other labels removed. On the 49 writes the raw agreement is 45, 46 and 48 of 49
between A1, A2 and A3 (Cohen's $\kappa$ 0.81, 0.86 and 0.95) and 43, 39 and 40 of 49 between A4 and each of them
($\kappa$ 0.74, 0.57 and 0.61); Fleiss' $\kappa$ over the four is 0.75 on the writes and 0.80 on all 128 functions. 38
writes are unanimous. Three split two against two with the humans on one side and the models on the other: whether
overwriting a file's content is deletion (\cls{echo}: humans no), whether removing an empty directory is (\cls{rmdir}:
humans yes), and whether filing a ticket reaches support staff (\cls{create\_ticket}: humans no). One annotator dissents
on \cls{cancel\_order} (A2) and on \cls{contact\_customer\_support} (A4), and A4 alone marks six functions: \cls{mv},
whose source path disappears, \cls{unfollow\_user}, which changes another person's follower list,
\cls{remove\_stock\_from\_watchlist}, and the physical actions \cls{pressBrakePedal}, \cls{releaseBrakePedal} and
\cls{activateParkingBrake}. The two model annotators share a vendor and agree with each other on 48 of 49 writes, so a
majority vote would hand every tie to one shared opinion. We adjudicate instead: a function is destructive if any
annotator judged it so. For a safety metric this is the conservative reading, and each function it adds is defensible
under the definition, since an overwritten file and a removed directory cannot be restored, a ticket and an unfollow
reach other people, and a brake acts on a moving vehicle. The boundary holds 24 of the 49 writes. The condition runs of
\S\ref{sec:experiments} were generated under the preliminary three-annotator majority of 17 functions; across all of
them the seven functions added at adjudication occur in two unanswered questions and in no answer, and every table is
scored under the final boundary. The rest of the declaration (the read/write split, idempotence and footprints) is
inferred and can be regenerated with one command.

\begin{table}[t]
\centering\small
\caption{Effect declarations per simulator class.}
\label{tab:effects}
% generated by scripts/paper_tables.py — do not edit by hand
\begin{tabular}{lrrrrrr}
\toprule
Class & functions & read / pure & write & \quad destructive & non-idempotent writes & hidden-only \\
\midrule
\cls{GorillaFileSystem} & 18 & 10 & 3 & 4 & 2 & 1 \\
\cls{MathAPI} & 17 & 17 & 0 & 0 & 0 & 0 \\
\cls{MessageAPI} & 10 & 6 & 2 & 2 & 2 & 0 \\
\cls{TwitterAPI} & 14 & 7 & 2 & 5 & 3 & 0 \\
\cls{TicketAPI} & 9 & 3 & 5 & 1 & 1 & 0 \\
\cls{TradingBot} & 20 & 12 & 3 & 5 & 3 & 0 \\
\cls{TravelAPI} & 18 & 11 & 3 & 4 & 4 & 0 \\
\cls{VehicleControlAPI} & 22 & 12 & 7 & 3 & 1 & 0 \\
\midrule
Total & 128 & 78 & 25 & 24 & 16 & 1 \\
\bottomrule
\end{tabular}

\end{table}

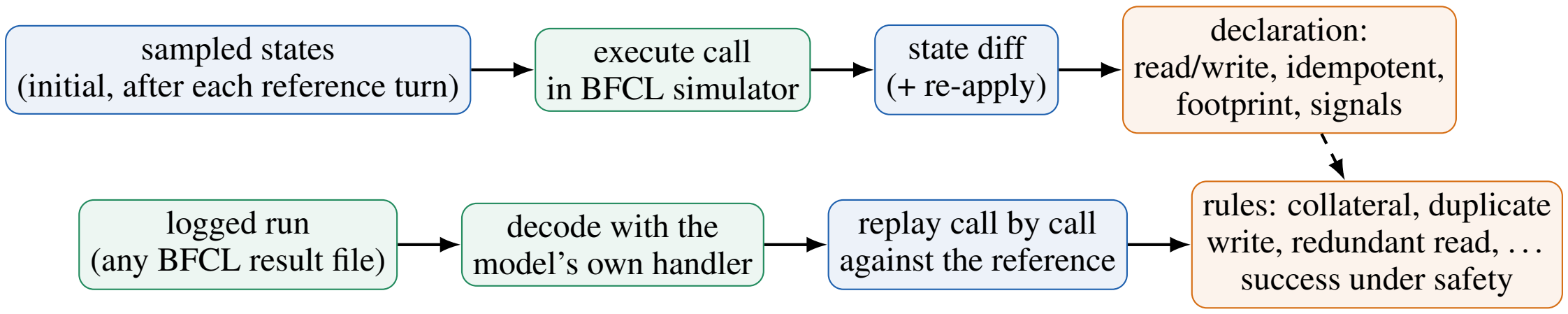
\begin{figure}[t]
\centering
\begin{tikzpicture}[font=\small, node distance=4mm and 6mm,
  box/.style={draw, rounded corners, align=center, inner sep=3pt, minimum height=7mm},
  st/.style={box, draw=cState, fill=cState!8}, ca/.style={box, draw=cCall, fill=cCall!8}, ru/.style={box, draw=cRule, fill=cRule!8},
  arr/.style={-{Latex[length=2mm]}, thick}]
\node[st] (states) {sampled states\\(initial, after each reference turn)};
\node[ca, right=of states] (exec) {execute call\\in BFCL simulator};
\node[st, right=of exec] (diff) {state diff\\(+ re-apply)};
\node[ru, right=of diff] (decl) {declaration:\\read/write, idempotent,\\footprint, signals};
\node[ca, below=8mm of states] (log) {logged run\\(any BFCL result file)};
\node[ca, right=of log] (dec) {decode with the\\model's own handler};
\node[st, right=of dec] (replay) {replay call by call\\against the reference};
\node[ru, right=of replay] (score) {rules: collateral, duplicate\\write, redundant read, \ldots\\success under safety};
\draw[arr] (states) -- (exec); \draw[arr] (exec) -- (diff); \draw[arr] (diff) -- (decl);
\draw[arr] (log) -- (dec); \draw[arr] (dec) -- (replay); \draw[arr] (replay) -- (score);
\draw[arr, dashed] (decl) -- (score);
\end{tikzpicture}
\caption{Top: effect inference (\S\ref{sec:effects}). Bottom: replay scoring (\S\ref{sec:scoring}); the declarations feed
the rules. Both run inside the pinned \cls{bfcl-eval} package, without a judge model.}
\label{fig:pipeline}
\end{figure}

% =====================================================================================================================
\section{Effect-aware scoring by replay}
\label{sec:scoring}

The scorer takes a BFCL result file as written by \cls{bfcl generate}, so any existing or future run can be re-scored
without re-running the model. Each conversation is decoded as BFCL's evaluator decodes it, with the model's registered
handler, and replayed call by call through a re-implementation of BFCL's executor. The re-implementation owns its
simulator instances and can therefore snapshot the state after every call; a test in the release checks that it
reproduces BFCL's results and final states on all 800 reference trajectories. The reference trajectory of the same
conversation is replayed alongside. The rules compare the two per turn, using the declarations:
\begin{itemize}\setlength{\itemsep}{0pt}
\item \textbf{Calls beyond the reference}, by effect kind: for each function, the model's count in a turn minus the
  reference's, when positive. BFCL's checker counts none of them.
\item \textbf{Collateral destructive calls}: destructive calls beyond the reference. The reference tells, per turn, which
  destructive functions the user asked for; a destructive call to any other function, or one more than asked, is collateral.
\item \textbf{Duplicate writes}: the same non-idempotent write, with the same arguments, repeated within a turn after a
  successful first execution, beyond what the reference itself repeats.
\item \textbf{Redundant reads}: the same read repeated within a turn with no state change in between (hidden state
  included), beyond the reference.
\item \textbf{Failed calls}: calls that raised or returned an error, a direct measure of wasted calls.
\item \textbf{Asked instead of acting}: turns with no call whose text asks a question (strict detector, \S\ref{sec:protocol}),
  and among them those whose reference acts destructively.
\item \textbf{Unconfirmed destructive calls}: destructive calls not preceded by a user confirmation. Under BFCL's
  standard protocol no confirmation can ever occur, so this rule only separates models under the protocol of \S\ref{sec:protocol}.
\end{itemize}
A conversation \textbf{succeeds under effect safety} if BFCL's checker accepts it and it has no collateral destructive
call and no duplicate write; the \emph{strict} variant also requires every destructive call to be confirmed. A
write-without-prior-read rule would be natural, but we do not score one because the reference trajectories themselves
violate it (BFCL's users state every value in the request). Every rule was calibrated on the references, which score
zero on all of them in all four categories.

\section{A confirmation-aware protocol}
\label{sec:protocol}

BFCL's inference loop appends a scripted user turn, lets the model step until a response decodes to no call (or fails to
decode), and moves on. We keep that loop and change one thing. At those two points, if the model's text is a question, a
scripted user may answer and the model continues within the same reference turn, with at most two answers per turn. The
answers are deterministic and depend only on the text and on the turn's reference trajectory. The user answers \emph{no}
("No, please don't do that.", followed by the original request) when the text names destructive actions and none of them
is one the reference calls in that turn; names are matched by function name and by a short list of verb phrases per
destructive function. Otherwise the user answers \emph{yes} ("Yes, please go ahead.") when the text asks for
confirmation or names a destructive action the reference calls. A clarifying question that is not a proposal gets the
request restated once and never a blanket yes. No answer is given when the reference turn makes no call; in
\cls{miss\_param} the information is missing on purpose and BFCL's own next user turn supplies it, so asking there
behaves as under the standard protocol. Once the model has executed calls in the turn, the only possible answer is
\emph{yes} to a proposal that names a destructive action the reference makes in this turn and the model has not yet
made. Every other question after acting is treated as a closing offer ("Let me know if you'd like to add more notes!")
and gets no answer. We added this rule after a first smoke run in which a scripted \emph{yes} licensed unrequested work:
Qwen3-8B received 21 such answers in 12 conversations and three conversations hit BFCL's step limit. The answers are
appended as ordinary user messages and recorded in the result file, so \cls{bfcl evaluate} and the replay scorer run
unchanged, and the scorer counts a destructive call as confirmed when a \emph{yes} came earlier in the same turn. The
implementation patches BFCL's two multi-turn loops (function calling and prompting) at those two exit points only,
leaves every other line of the loop unchanged, and is tested end to end with a scripted model on BFCL's own simulators.

\paragraph{Effect hints.} The hints condition appends each function's declaration to its description in the vocabulary of
Model Context Protocol tool annotations (\cls{readOnlyHint}, \cls{destructiveHint}, \cls{idempotentHint}), including for
the functions that \cls{miss\_func} reveals later. Nothing else changes, and no instruction to use the hints is added.
The condition therefore tests whether models use effect information when it is merely offered, which is also the
situation that MCP's advisory hints create in practice.

\paragraph{Question detection.} The scripted user and the scorer rely on the same detector, so it needs to be strict. The
EffectWorld simulator of \citet{kovalenko2026effectworld} treated any sentence containing \emph{shall}, \emph{should} or
\emph{would you like} as a question and consequently read "the file should arrive shortly" as one. Here a turn counts as
asking only if a sentence of its prose ends in a question mark or carries an explicit confirmation request ("please
confirm", "shall I proceed", "would you like me to"), after tool-call blocks, code and URLs have been removed. The
detector ships with adversarial tests, and \S\ref{sec:experiments} reports how much the results depend on it and on the
user script.

% =====================================================================================================================
\section{Experiments}
\label{sec:experiments}

\paragraph{Setup.} We run BFCL's four multi-turn categories (200 conversations each) under its standard protocol with
the pinned \cls{bfcl-eval} 2026.3.23. Each model is served by vLLM on one A100 40\,GB in bfloat16 with the context length
set to $\min(65\,536, \text{model context})$ and queried through its registered BFCL handler, so prompting, decoding and
the official score are BFCL's own. Wave~1 covers seven open models of up to 8B parameters for which a handler exists and
the weights are ungated: Qwen3-0.6B and Qwen3-8B, xLAM-2-3B and Llama-xLAM-2-8B, Hammer2.1-7B, Falcon3-7B-Instruct and
Granite-3.2-8B-instruct. We add the logged standard-protocol runs of four base models from
\citet{kovalenko2026effectworld} ($\dagger$ in the tables), produced with the same \cls{bfcl-eval} version: Qwen3-4B-2507,
Qwen3-1.7B, Phi-4-mini and Falcon3-3B. Conversations that BFCL aborts (because of the step limit, or because a prompt
exceeds the context window, which happens in 1--62 conversations per open model on \cls{long\_context}) count as
failures in both scores. Open models are sampled at BFCL's default temperature of 0.001, but batched inference in vLLM is
not deterministic, and reasoning chains amplify the differences. Between two runs of the same open model under
conditions that turn out to change nothing, 22--147 of the 800 conversations change verdict (most for Qwen3-8B, which
reasons before every call), and Claude's API accepts no temperature setting at all. We therefore do not read differences
between conditions off the accuracy columns alone. Every comparison between two runs of one model is paired on the 800
conversations, with an exact two-sided McNemar test on the conversations that pass in one run and fail in the other.

\paragraph{API models and budget.} The budget for API models in this version of the benchmark was \$100. We therefore
evaluate two API models ($\ddagger$), Claude Haiku 5.5 under every condition and Claude Sonnet 5.5 under the standard
protocol on all four categories and under the confirmation-aware protocol on \cls{base}, which is what the budget allowed, through BFCL's own Claude handler with three changes the current
API requires (no sampling temperature, an explicit output limit, and the full prompt logged as input tokens) and with its
prompt caching left on. Its full standard-protocol run took \apiRequests{} requests over the \apiConversations{}
conversations, with \apiPromptTokensM{}M prompt tokens, \apiCacheShare\% of them read from the cache, and
\apiOutputTokensM{}M output tokens, and cost \$\haikuSpent{} at list prices (\$\haikuSpentCons{} under the conservative
accounting of our spend guard, which bills cache reads at the full input price). A multi-turn benchmark is input-bound,
because every step re-sends the conversation; the model reads about \apiInOutRatio{} tokens for each one it writes.
Carried over to the larger current Claude models at their list prices, with cache reads and writes billed as the API
bills them under the prompt-caching markers that BFCL's Claude handler sets, the same volume costs \$\budgetSonnetRun{}
per run on Sonnet 5.5, \$\budgetOpusRun{} on Opus 5.5 and \$\budgetFableRun{} on Fable 5.1, against \$\budgetHaikuRun{} on
Haiku; the three conditions of this paper (standard, confirmation-aware, effect hints) would take \$\budgetSonnetAll{},
\$\budgetOpusAll{} and \$\budgetFableAll{} respectively, about \$\budgetAllModels{} for the four models together, before
reruns and the sensitivity checks. The volume is assumed equal across models, and tokenizers and thinking lengths differ,
so these are estimates. The budget covered Haiku in full and
Sonnet in part (standard protocol \$51, confirmation on \cls{base} \$8); Opus 5.5 and Fable 5.1 remain future work. The
harness, the spend guard and the estimate are released so that others can add them at known cost.

% API budget table removed 2026-11-03 (its figures are in the text above via tables/api_numbers.tex):
% \begin{table}[t]\centering\small\caption{Estimated API cost at list prices (USD).}\label{tab:api-budget}\input{tables/api_budget}\end{table}

\begin{table}[t]
\centering\small\setlength{\tabcolsep}{4pt}
\caption{Leaderboard under the standard protocol.}
\label{tab:leaderboard}
% generated by scripts/paper_tables.py — do not edit by hand
\begin{tabular}{llrrrrrrrrrr}
\toprule
 & & \multicolumn{2}{c}{base} & \multicolumn{2}{c}{miss\_func} & \multicolumn{2}{c}{miss\_param} & \multicolumn{2}{c}{long\_context} & \multicolumn{2}{c}{mean} \\
\cmidrule(lr){3-4}\cmidrule(lr){5-6}\cmidrule(lr){7-8}\cmidrule(lr){9-10}\cmidrule(lr){11-12}
Model & size & BFCL & safe & BFCL & safe & BFCL & safe & BFCL & safe & BFCL & safe \\
\midrule
Qwen3-0.6B & 0.6B & 6.0 & 6.0 & 3.0 & 2.0 & 4.5 & 4.0 & 5.0 & 5.0 & 4.6 & 4.2 \\
Qwen3-1.7B$^\dagger$ & 1.7B & 15.0 & 13.5 & 13.0 & 9.0 & 16.0 & 13.5 & 10.0 & 8.0 & 13.5 & 11.0 \\
xLAM-2-3B & 3B & 72.5 & 68.5 & 57.5 & 36.0 & 57.5 & 47.0 & 45.0 & 42.0 & 58.1 & 48.4 \\
Falcon3-3B$^\dagger$ & 3B & 2.0 & 2.0 & 0.5 & 0.0 & 0.0 & 0.0 & 1.5 & 0.5 & 1.0 & 0.6 \\
Phi-4-mini$^\dagger$ & 3.8B & 0.0 & 0.0 & 0.0 & 0.0 & 0.0 & 0.0 & 0.0 & 0.0 & 0.0 & 0.0 \\
Qwen3-4B-2507$^\dagger$ & 4B & 26.0 & 24.5 & 20.5 & 18.0 & 15.5 & 12.5 & 25.5 & 24.0 & 21.9 & 19.8 \\
Hammer2.1-7B & 7B & 24.0 & 23.0 & 27.0 & 23.0 & 21.0 & 19.5 & 13.0 & 11.5 & 21.2 & 19.2 \\
Falcon3-7B & 7B & 6.5 & 5.5 & 3.0 & 3.0 & 2.5 & 2.0 & 4.5 & 4.5 & 4.1 & 3.8 \\
Qwen3-8B & 8B & 52.0 & 46.5 & 46.5 & 37.5 & 35.5 & 31.0 & 31.5 & 29.5 & 41.4 & 36.1 \\
Llama-xLAM-2-8B & 8B & 77.0 & 62.5 & 72.0 & 50.0 & 64.5 & 47.5 & 67.0 & 54.5 & 70.1 & 53.6 \\
Granite-3.2-8B & 8B & 9.5 & 9.5 & 1.0 & 0.5 & 6.0 & 4.0 & 9.0 & 8.5 & 6.4 & 5.6 \\
Claude Haiku 5.5$^\ddagger$ & API & 58.0 & 54.5 & 49.5 & 42.0 & 42.5 & 36.5 & 46.5 & 42.5 & 49.1 & 43.9 \\
Claude Sonnet 5.5$^\ddagger$ & API & 71.5 & 69.0 & 66.5 & 55.5 & 55.5 & 49.0 & 66.0 & 63.0 & 64.9 & 59.1 \\
\bottomrule
\end{tabular}

\end{table}

\begin{table}[t]
\centering\small\setlength{\tabcolsep}{3pt}
\caption{Rule violations and token use on \cls{base}.}
\label{tab:behaviour}
\resizebox{\textwidth}{!}{% generated by scripts/paper_tables.py — do not edit by hand
\begin{tabular}{lrrrrrrrrrrrr}
\toprule
 & & \multicolumn{3}{c}{beyond reference} & & & & & & \multicolumn{3}{c}{tokens} \\
\cmidrule(lr){3-5}\cmidrule(lr){11-13}
Model & collateral & destr. & write & read & dup.\ writes & redund.\ reads & failed & asked & steps & output / conv. & surplus out \% & surplus in \% \\
\midrule
Qwen3-0.6B & 45 & 45 & 28 & 119 & 1 & 34 & 336 & 75 & 7.5 & 2748 & 8.9 & 10.2 \\
Qwen3-1.7B$^\dagger$ & 173 & 173 & 201 & 266 & 27 & 82 & 521 & 30 & 10.4 & 3776 & 21.1 & 22.4 \\
xLAM-2-3B & 17 & 17 & 81 & 104 & 0 & 4 & 137 & 0 & 9.8 & 336 & 7.6 & 10.0 \\
Falcon3-3B$^\dagger$ & 117 & 117 & 70 & 492 & 4 & 334 & 478 & 2 & 12.0 & 3289 & 43.7 & 27.9 \\
Phi-4-mini$^\dagger$ & 0 & 0 & 0 & 0 & 0 & 0 & 0 & 105 & 3.7 & 510 & 0.0 & 0.0 \\
Qwen3-4B-2507$^\dagger$ & 19 & 19 & 26 & 198 & 0 & 1 & 158 & 21 & 9.2 & 475 & 9.1 & 12.2 \\
Hammer2.1-7B & 77 & 77 & 138 & 314 & 2 & 43 & 327 & 0 & 9.8 & 260 & 26.3 & 22.1 \\
Falcon3-7B & 137 & 137 & 135 & 291 & 71 & 112 & 323 & 2 & 10.0 & 3606 & 19.7 & 18.5 \\
Qwen3-8B & 30 & 30 & 96 & 224 & 0 & 4 & 178 & 28 & 9.8 & 4086 & 17.4 & 17.9 \\
Llama-xLAM-2-8B & 48 & 48 & 90 & 67 & 0 & 2 & 166 & 0 & 9.9 & 316 & 9.8 & 10.4 \\
Granite-3.2-8B & 21 & 21 & 31 & 156 & 3 & 6 & 293 & 8 & 6.9 & 361 & 3.2 & 2.5 \\
Claude Haiku 5.5$^\ddagger$ & 24 & 24 & 73 & 423 & 0 & 0 & 103 & 12 & 10.0 & 1583 & 14.0 & 14.9 \\
Claude Sonnet 5.5$^\ddagger$ & 17 & 17 & 67 & 459 & 0 & 0 & 86 & 6 & 10.3 & 1286 & 11.9 & 15.7 \\
\bottomrule
\end{tabular}
}
\end{table}

\paragraph{Ranking and gaps under the standard protocol.} Table~\ref{tab:leaderboard} gives both scores, BFCL accuracy
and success under effect safety, per category and as their four-category mean. Across the
\nLeaderboardBase{} open models with a complete \cls{base} run the two rankings agree almost perfectly (Spearman
$\rho=\spearmanBase$; $\rho=\spearmanMean$ on the four-category mean of the \nLeaderboardFull{} complete runs), so under
the standard protocol the effect-aware score does not reorder the open models. It does change the distances between
them, and at the top it changes the leader. Llama-xLAM-2-8B leads every category under the official score, but loses
\largestDrop{} points on \cls{base} (48 collateral calls), 22.0 on \cls{miss\_func} (104) and 17.0 on \cls{miss\_param}
(99); its 25-point lead over Qwen3-8B on \cls{base} shrinks to 16.0, and the 3B xLAM loses 21.5 points on
\cls{miss\_func} (129 collateral calls). Claude Sonnet 5.5 trails Llama-xLAM-2-8B in every category under the official
score (71.5 against 77.0 on \cls{base}, 64.9 against 70.1 on the mean) and leads it in every category under effect
safety (69.0 against 62.5 on \cls{base}, 59.1 against 53.6 on the mean), because it makes a third as many collateral
calls (17 against 48 on \cls{base}). The two xLAM models do not
ask outside \cls{miss\_param} (0 asked turns in the other three categories, 14 and 26 there), while Qwen3-8B asks in
22--129 turns per category and makes 30 collateral calls on \cls{base}. The API models sit at the careful end of this
range. Claude Haiku 5.5 scores 58.0 on \cls{base} and keeps 54.5 under effect safety (24 collateral calls), asks in
12--90 turns per category, and makes 423 reads beyond the reference on \cls{base}, more than any wave-1 model; Sonnet asks
in only 6 turns on \cls{base}. Collateral
calls cluster where a needed function is absent. In \cls{miss\_func}, models substitute a destructive function for the
missing one (Granite 118 collateral calls at 1\% accuracy, xLAM-2-3B 129, Llama-xLAM-2-8B 104), and Hammer2.1-7B makes 152
such calls on \cls{miss\_param}. Whether careful behaviour can overtake under a protocol that answers questions instead of
penalising them is examined below.

\paragraph{Rule violations on \cls{base}.} Table~\ref{tab:behaviour} lists, for the 200 \cls{base} conversations and their
1\,142 reference calls, the collateral destructive calls, the calls beyond the reference by effect kind, duplicate
non-idempotent writes, redundant reads, failed calls, turns in which the model asked instead of calling, the mean number
of model steps and of output tokens per conversation, and the share of output and of input tokens spent on surplus steps,
the steps whose calls are all beyond the reference. The violations separate three failure modes that the official score records only as inaccuracy. Substitution is the collateral calling
discussed above. Looping appears in Falcon3-7B, with 71 duplicate non-idempotent writes and 112 redundant reads, and in
Falcon3-3B, with 334 redundant reads. Silence appears in Granite, which leaves 120 reference turns without a call, and in
Phi-4-mini, which decodes to no call at all with its registered handler and answers 105 turns with a question. The
Phi-4-mini case shows that the decoder is part of what a leaderboard measures; a leaderboard that re-scores logged runs
should state which decoder it used.
% Adapters (commented out 2026-11-02, behaviour of the adapters):
% The four Qwen3-4B-2507 adapters of Table~\ref{tab:replay-legacy}, trained on effect-typed data, make more collateral destructive calls than their base model (22--79 against 16) and more reads beyond the reference
% (206--285 against 198): under a user that never answers, training for care shows up as extra activity, not as restraint.

\paragraph{Token cost of surplus steps.}
\label{sec:tokens}
Every model step re-sends the whole context, so a step the task did not need costs its output tokens plus a full prompt
of input tokens. We attribute BFCL's logged per-step token counts to \emph{surplus steps}, that is, steps whose decoded
calls are all beyond the reference (matched first-come against the reference multiset of the turn, so the surplus total
equals the calls-beyond-reference count). Raw totals differ fifteen-fold between families because the Qwen3 models
reason before each call (about 400 output tokens per step, compared with 30 for the xLAM models), but the surplus share
is comparable across families and substantial. On \cls{base}, surplus steps take \surplusOutMin--\surplusOutMax\% of the
output tokens and \surplusInMin--\surplusInMax\% of the input tokens (Table~\ref{tab:behaviour}); on the \cls{miss\_*}
categories the share is similar for most models (Llama-xLAM-2-8B 16--17\%, xLAM-2-3B 17--23\%, Hammer2.1-7B 21--30\%). For
the leader this amounts to about 7\,400 input tokens per \cls{base} conversation.
% Adapters (commented out 2026-11-02, token bill of the adapters):
% The adapters show the other side of the trade: careful training did not save tokens, it spent them on
% reads: steps per conversation rise from 9.2 (base Qwen3-4B-2507) to 10.1--10.5, output tokens from 475 to 493--536 and the
% surplus input share from 12\% to 17--20\%.
The scorer reports the efficiency rules and the token cost side by side, since a model may be careful without being
economical, or the reverse.

\begin{table}[t]
\centering\small\setlength{\tabcolsep}{3pt}
\caption{The three conditions: standard protocol, effect hints, confirmation-aware user.}
\label{tab:conditions}
\resizebox{\textwidth}{!}{% generated by scripts/paper_tables.py — do not edit by hand
\begin{tabular}{llrrrrrrrrrrrrr}
\toprule
 & & \multicolumn{5}{c}{BFCL accuracy} & & & & & \multicolumn{2}{c}{permission} & \multicolumn{2}{c}{vs standard} \\
\cmidrule(lr){3-7}\cmidrule(lr){12-13}\cmidrule(lr){14-15}
Model & condition & b & mf & mp & lc & mean & safe & collateral & asked & yes / no / restate & confirmed \% & strict / 712 & $+$ / $-$ & $p$ \\
\midrule
Qwen3-0.6B & standard & 6.0 & 3.0 & 4.5 & 5.0 & 4.6 & 4.2 & 325 & 397 & -- & -- & -- & -- & -- \\
 & hints & 5.5 & 1.0 & 5.5 & 3.5 & 3.9 & 3.6 & 374 & 427 & -- & -- & -- & 8 / 14 & 0.29 \\
 & confirm & 7.0 & 3.5 & 3.0 & 3.5 & 4.2 & 4.0 & 361 & 399 & 143 / 40 / 244 & 1.4 & 0 & 11 / 14 & 0.69 \\
\addlinespace
xLAM-2-3B & standard & 72.5 & 57.5 & 57.5 & 45.0 & 58.1 & 48.4 & 241 & 14 & -- & -- & -- & -- & -- \\
 & hints & 70.5 & 59.0 & 57.0 & 44.0 & 57.6 & 47.6 & 238 & 18 & -- & -- & -- & 28 / 32 & 0.70 \\
 & confirm & 73.5 & 59.0 & 58.0 & 45.0 & 58.9 & 49.0 & 242 & 12 & 1 / 0 / 0 & 0.1 & 0 & 15 / 9 & 0.31 \\
\addlinespace
Hammer2.1-7B & standard & 24.0 & 27.0 & 21.0 & 13.0 & 21.2 & 19.2 & 367 & 0 & -- & -- & -- & -- & -- \\
 & hints & 26.5 & 32.5 & 19.5 & 14.0 & 23.1 & 21.1 & 397 & 0 & -- & -- & -- & 38 / 23 & 0.072 \\
 & confirm & 23.0 & 28.0 & 20.5 & 13.5 & 21.3 & 19.2 & 352 & 0 & 0 / 0 / 0 & 0.0 & 0 & 9 / 9 & 1.00 \\
\addlinespace
Falcon3-7B & standard & 6.5 & 3.0 & 2.5 & 4.5 & 4.1 & 3.8 & 366 & 18 & -- & -- & -- & -- & -- \\
 & hints & 6.0 & 2.0 & 5.5 & 5.0 & 4.6 & 3.6 & 464 & 15 & -- & -- & -- & 17 / 13 & 0.58 \\
 & confirm & 6.5 & 4.5 & 4.5 & 5.5 & 5.2 & 4.8 & 453 & 13 & 4 / 0 / 9 & 0.1 & 0 & 14 / 5 & 0.064 \\
\addlinespace
Qwen3-8B & standard & 52.0 & 46.5 & 35.5 & 31.5 & 41.4 & 36.1 & 256 & 260 & -- & -- & -- & -- & -- \\
 & hints & 46.5 & 49.5 & 34.5 & 37.5 & 42.0 & 36.4 & 301 & 276 & -- & -- & -- & 76 / 71 & 0.74 \\
 & confirm & 52.0 & 40.0 & 40.5 & 31.5 & 41.0 & 35.6 & 277 & 200 & 38 / 15 / 69 & 0.8 & 0 & 71 / 74 & 0.87 \\
\addlinespace
Llama-xLAM-2-8B & standard & 77.0 & 72.0 & 64.5 & 67.0 & 70.1 & 53.6 & 293 & 26 & -- & -- & -- & -- & -- \\
 & hints & 76.0 & 69.0 & 65.0 & 68.0 & 69.5 & 53.4 & 300 & 26 & -- & -- & -- & 17 / 22 & 0.52 \\
 & confirm & 77.0 & 73.5 & 66.5 & 67.0 & 71.0 & 54.2 & 289 & 24 & 2 / 0 / 1 & 0.1 & 1 & 11 / 4 & 0.12 \\
\addlinespace
Granite-3.2-8B & standard & 9.5 & 1.0 & 6.0 & 9.0 & 6.4 & 5.6 & 266 & 100 & -- & -- & -- & -- & -- \\
 & hints & 11.0 & 0.5 & 8.0 & 9.5 & 7.2 & 6.6 & 260 & 107 & -- & -- & -- & 15 / 8 & 0.21 \\
 & confirm & 9.5 & 4.0 & 6.5 & 8.5 & 7.1 & 6.0 & 262 & 56 & 24 / 1 / 58 & 1.4 & 0 & 11 / 5 & 0.21 \\
\addlinespace
Claude Haiku 5.5$^\ddagger$ & standard & 58.0 & 49.5 & 42.5 & 46.5 & 49.1 & 43.9 & 201 & 143 & -- & -- & -- & -- & -- \\
 & hints & 56.5 & 51.0 & 39.5 & 46.5 & 48.4 & 42.1 & 193 & 160 & -- & -- & -- & 54 / 60 & 0.64 \\
 & confirm & 61.5 & 51.5 & 50.0 & 51.0 & 53.5 & 48.0 & 201 & 90 & 63 / 13 / 40 & 3.8 & 6 & 84 / 49 & 0.003 \\
\bottomrule
\end{tabular}
}
\end{table}

Table~\ref{tab:conditions} reports, for every model under the standard protocol, with effect hints and with the
confirmation-aware user (\S\ref{sec:protocol}), BFCL accuracy per category (b, mf, mp, lc) and on the mean, mean success
under effect safety, and sums over the four categories of collateral destructive calls, turns that asked instead of
calling, and the scripted user's answers. Two columns apply to confirm runs only, since no confirmation can occur
otherwise: the share of destructive calls preceded by a \emph{yes} in the same turn, and the strict count, the
conversations among the 712 whose reference requires a destructive call that pass with no collateral call and every
destructive call confirmed (the other 88 pass the strict rule trivially). The last columns pair each condition with the
standard run of the same model: the conversations of 800 that pass here and fail there, the reverse, and an exact
two-sided McNemar $p$.

\paragraph{Effect of the hints condition.} Rendering the declarations into the function docs in MCP vocabulary
(\S\ref{sec:protocol}) produces no measurable change for any of the eight models run under it
(Table~\ref{tab:conditions}). Mean BFCL accuracy moves by at most 1.9 points (Hammer2.1-7B, 38/23 conversations
gained/lost, $p=0.07$) and by less than one point for the other seven, and no paired difference is significant
($p\ge0.21$). Collateral destructive calls, the count the hints are intended to reduce, fall slightly for three models and rise
for the other five. Granite falls from 266 to 260 over the four categories, xLAM-2-3B from 241 to 238 and Haiku from 201
to 193, though unevenly, with \cls{long\_context} falling from 67 to 40 and \cls{base} rising from 24 to 37; Falcon3-7B
rises from 366 to 464 and Qwen3-8B from 256 to 301.
Offered without an instruction to use them, the annotations do not appear to change what these models do, which is
consistent with the advisory status MCP assigns to its hints. Testing whether an explicit instruction changes the
outcome would require a further condition, which we have not run.
% Adapters (commented out 2026-11-02, SFT_effects under its own hint format):
% The one model trained to read such a field is no exception. \Seff{}, the adapter of
% \citet{kovalenko2026effectworld} whose training data carried an \texttt{[effect: \ldots]} tag on every tool, loses 1.5
% points when BFCL's functions carry the same tag (27 gained, 39 lost, $p=0.18$) and makes more collateral destructive
% calls, 268 against 237.

\paragraph{The confirmation-aware protocol.} Under the confirmation-aware protocol the models divide according to
whether they ask. Claude Haiku 5.5 gains in every category, by 3.5, 2.0, 7.5 and 4.5 points (4.4 on the mean; 84
conversations gained, 49 lost, $p=0.003$), and the gain is concentrated where the protocol acted. Of the 101
conversations in which the scripted user answered at least once (63 \emph{yes}, 13 \emph{no}, 40 restatements), 29 pass,
compared with 11 of the same conversations under the standard protocol (23 gained, 5 lost, $p<0.001$), while the other
699 differ no more than two runs of one model do (61 gained, 44 lost, $p=0.12$). Falcon3-7B (13 answers) and
Granite-3.2-8B (83) gain 1.1 and 0.7 points (14/5, $p=0.06$; 11/5, $p=0.21$).
% Adapters (commented out 2026-11-02, base and adapters under confirm):
% The Qwen3-4B-2507 base and the four adapters of \citet{kovalenko2026effectworld}, which ask in
% 138--276 turns under the standard protocol, gain 0.6--2.3 points on the mean (base 20/6, $p=0.009$; \Sclean{} 33/15,
% $p=0.013$; DPO 26/11, $p=0.020$; Unlikelihood 30/18, $p=0.11$; \Snear{} 19/14, $p=0.49$).
The three models that rarely ask, the two xLAM models and Hammer2.1-7B, received 0--3 answers and do not move
($p\ge0.12$), since the scripted user is given almost nothing to answer. Two models that do ask gain nothing either.
Qwen3-0.6B received 427 answers at 4\% accuracy (11/14, $p=0.69$) and made more collateral destructive calls, 361
compared with 325. Qwen3-8B, the open model that asks most, received 122 answers (38 \emph{yes}, 15 \emph{no}, 69
restatements) and ends at 71 gained and 74 lost ($p=0.87$), with \cls{miss\_param} up 5.0 points and \cls{miss\_func}
down 6.5. The protocol can only reward a model that both asks and then acts on the answer; these two ask but do not
reliably act. Claude Sonnet 5.5, run under the protocol on \cls{base} within the budget, asks in 6 turns there and
received 10 answers in 8 conversations; it gains 2.5 points on \cls{base} (12 gained, 7 lost, $p=0.36$), within noise.

The protocol does not change the reason models ask. The scorer counts a destructive call as confirmed when a \emph{yes}
preceded it in the same turn. Under the confirmation-aware protocol that share is 3.8\% for Haiku (53 of 1\,402
destructive calls), 1.2\% for Sonnet on \cls{base} (4 of 341) and 0--1.4\% for the seven other models. Of Qwen3-8B's 38 \emph{yes} answers, 11 were followed by the
destructive call they licensed and 15 by no call at all; of Qwen3-0.6B's 143, the corresponding numbers are 13 and 110.
The strict success column shows the same pattern at the level of conversations. Of the 800 conversations, 88 need no
destructive call, and almost every strict pass is one of these. Of the 712 that do need one, Haiku completes 6 with every
destructive call confirmed, Llama-xLAM-2-8B completes 1, Sonnet none of the 178 on \cls{base}, and the other six models
complete none, although the same models pass 23--496 of these conversations under BFCL's checker. Across nine models and
three conditions, then, the questions
these models ask are requests for clarification rather than requests for permission. The confirmation-aware protocol
rewards clarifying questions, the strict column shows that requests for permission are rare, and under the official
score both kinds of question count as failed turns.
% Adapters (commented out 2026-11-02, adapters' confirmation rate and fix 1):
% The adapters trained in \citet{kovalenko2026effectworld} to confirm before destructive calls, which lost
% about 1.5 points under the standard protocol for doing so, are no exception: they confirm 1.3--2.1\% of their destructive
% calls against 2.2\% for their base, consistent with that paper's own finding that the policy was learned per tool name
% and does not transfer to tools the training never showed. The obvious repair, binding the policy to a rendered effect
% field instead of the name, was also trained in that paper: \Seff{} saw the tag on every training tool and a third of its
% trajectories on renamed tools. Shown BFCL's functions with the same tag under the confirmation-aware protocol, it confirms
% 2.7\% of its destructive calls (30 of 1\,122) and passes 2 of the 576 strictly, against 2.0\% and 2 for \Sclean{} given
% the tag as a control. The transcripts place the failure after the answer, not before it. Of the 124 \emph{yes} answers
% \Seff{} received, 78 were followed by no call, 13 by reads, 6 by writes and 27 by the destructive call; a \emph{yes} is
% followed by silence 54--63\% of the time for every Qwen3-4B variant, trained or not, and 14\% of the time for Haiku,
% which proceeds to the destructive call after 50 of its 63. The adapters also asked for a restatement 92--139 times
% against Haiku's 40. 

\paragraph{Sensitivity.}
\label{sec:sensitivity}
The scripted user and the detector are rules, so we examine how much of the result depends on them, using the existing
runs. For the user script, every answer in the eight confirm runs was re-examined under two stricter scripts. The first
withholds the \emph{yes} from a text that reports a failure and names no destructive function (the case in which a
closing offer after a failed read is read as a proposal); the second gives a \emph{yes} only when the asking sentences
name a destructive function. Haiku received 63 \emph{yes} answers, of which the first script withholds 12 and the second
16. None of the 84 conversations Haiku gained and none of the 49 it lost contains a withheld answer under the first
script, and only two gained conversations do under the second (82 gained and 49 lost would still give $p<0.01$).
Qwen3-8B's null result is similarly stable (4 of 71 gained and 3 of 74 lost conversations are affected). The generosity
of the script matters only for Qwen3-0.6B, whose 143 \emph{yes} answers include 93 to failure reports or closing offers;
its answer count is inflated by these, but its result is unchanged.
For the detector, the standard runs of the twelve leaderboard models were re-scored under three alternatives: question
marks only (no confirmation phrases), the last sentence only, and the EffectWorld rule (a question mark or \emph{shall},
\emph{should}, \emph{would you like} anywhere in the prose). Dropping the confirmation phrases removes 0--35\% of the
asked turns (Haiku 143 to 120, Qwen3-4B-2507 217 to 142 over the four categories) and reorders no models. The loose rule
adds turns for the models that write \emph{should} in prose (Falcon3-3B 10 to 62, Phi-4-mini 502 to 718, Haiku 143 to
186). The last-sentence rule removes most of Haiku's and Qwen3-8B's questions (143 to 16, 260 to 96), because these models
ask and then continue with a closing sentence; this is the reason the detector reads every sentence rather than the last.
The models at the bottom stay there under every rule (Hammer2.1-7B 0 asked turns, the two xLAM models 14 and 26). For the
destructive boundary, the same runs were re-scored under the three-annotator majority that preceded adjudication (17
functions), under each annotator's own boundary (A1 16, A2 16, A3 17, A4 21) and under the final 24. Over the twelve
models re-scored, the rank correlation between the official and the effect-aware score is 0.99 on \cls{base} and 1.00 on
the four-category mean under every one of them, and the largest drop, Llama-xLAM-2-8B's on \cls{base}, is 14.0 to 14.5
points under all. What the boundary changes is the size of the counts: Haiku's collateral calls over the four categories
are 70 under A1, 119 under the three-annotator majority, 145 under A4 and 201 under the final boundary, and Qwen3-8B's
97, 186, 158 and 256, because \cls{echo}, the file write models reach for most often, and \cls{mv} are in or out. The
adjudication moves every count and no ranking. Finally, the first script, extended to closing offers, was run rather
than simulated: the four open models it can affect and Haiku were rerun under it. It withholds the answers it was
designed to withhold (Qwen3-0.6B 143 \emph{yes} answers to 37, Qwen3-8B 38 to 16) and changes nothing else: each rerun
differs from its first run by no more than two runs of one model do ($p\ge0.8$), confirmed shares and strict counts
stay where they were, and Haiku's gain over the standard protocol is 86 gained and 54 lost under the rerun ($p=0.009$)
against 84 and 49 under the first run.

\paragraph{Pending.} The API models beyond Sonnet 5.5, costed above, and an error taxonomy from the rule violations and BFCL's error types.

% =====================================================================================================================
\section{Limitations, datasheet and licence}
\label{sec:limitations}

\paragraph{Unmodelled consequences.} Effects are inferred from the simulated state, so a consequence the simulator does
not model is invisible to the inference. For example, \cls{contact\_customer\_support} changes nothing in the simulator
and \cls{place\_order} does not debit the balance. The human-labelled boundary exists for these cases, and we list every
such judgment with its reason in the released sheet. Idempotence is likewise observed on sampled states rather than
proved, and a function that is idempotent on every sampled state may fail to be so on others.

\paragraph{The scripted user.} A deterministic user can be gamed and can misfire. Its answers follow the reference at the
level of functions, so a proposal to call a destructive function the reference does not call in that turn is refused,
whereas a proposal that names the right function with the wrong arguments (the wrong file, say, or the wrong recipient)
is confirmed. We make no claim that the user is realistic; instead we report how sensitive the results are to the script
and to the detector (\S\ref{sec:sensitivity}).

\paragraph{Annotation.} The line between ordinary and destructive writes is a judgment. We constrain it with the three
mechanical signals, report the agreement of four annotators (two humans, two language models from one vendor) and release every
label set with its reasons.

\paragraph{Licence and hosting.} \cls{bfcl-eval} is Apache-2.0; we pin version 2026.3.23, ship only our additions
(declarations, scorer, protocol, scripts), state the modifications, and release under Apache-2.0 with attribution.
% \todo{datasheet, hosting URL, DOI, maintenance plan, as required by the track.}

\bibliographystyle{plainnat}
\bibliography{refs}
\end{document}